\pdfoutput=1

\documentclass[11pt]{article}

\usepackage[preprint]{acl}

\usepackage{times}
\usepackage{latexsym}
\usepackage{graphicx}

\usepackage[T1]{fontenc}

\usepackage[utf8]{inputenc}

\usepackage{microtype}

\usepackage{inconsolata}

\usepackage{listings}
\definecolor{codebg}{HTML}{F6F8FA}
\definecolor{codekw}{HTML}{0033B3}
\definecolor{codestr}{HTML}{067D17}
\definecolor{codecom}{HTML}{8C8C8C}
\lstdefinestyle{splytters}{
  language=Python,
  basicstyle=\ttfamily\small,
  backgroundcolor=\color{codebg},
  keywordstyle=\color{codekw}\bfseries,
  stringstyle=\color{codestr},
  commentstyle=\itshape\color{codecom},
  frame=single,
  rulecolor=\color{black!15},
  framesep=4pt,
  xleftmargin=5pt,
  xrightmargin=2pt,
  aboveskip=6pt,
  belowskip=2pt,
  breaklines=true,
  showstringspaces=false,
  columns=fullflexible,
  keepspaces=true,
}
\usepackage{booktabs}
\usepackage{tabularx}

\title{Splytters: A Python Library for Adversarial Data Splits}

\author{Stefan Larson \\
  Vanderbilt University \\
  Nashville, TN, USA \\\And
  Kirtan Shah \\
  AxiosPoint \\
  Toronto, ON, Canada \\}

\begin{document}
\maketitle
\begin{abstract}
Random train/test splits are the default practice in machine learning and NLP evaluation, yet they tend to place similar examples in both splits, which can overstate generalization.
Prior work has proposed \emph{non-random} splits, including \emph{adversarial} splits, which typically make the test set harder by making it dissimilar to the training set, but these methods are spread across papers and are rarely reusable across datasets or modalities.
We present \texttt{splytters}, an open-source Python library that implements non-random data splitting with a single, embedding-agnostic, scikit-learn-compatible interface.
The library provides 31 splitting algorithms for three objectives (\emph{adversarial}, \emph{overlap}, and \emph{balanced}) and \emph{sorters} that rank examples by difficulty, and it supports text, image, audio, and tabular data.
On various text, image, and audio tasks, adversarial splits can yield harder test sets than random splits and can thus be used to test generalizability.
\end{abstract}

\section{Introduction}
\label{sec:intro}

\begin{figure*}[t]
    \centering
    \includegraphics[width=\textwidth]{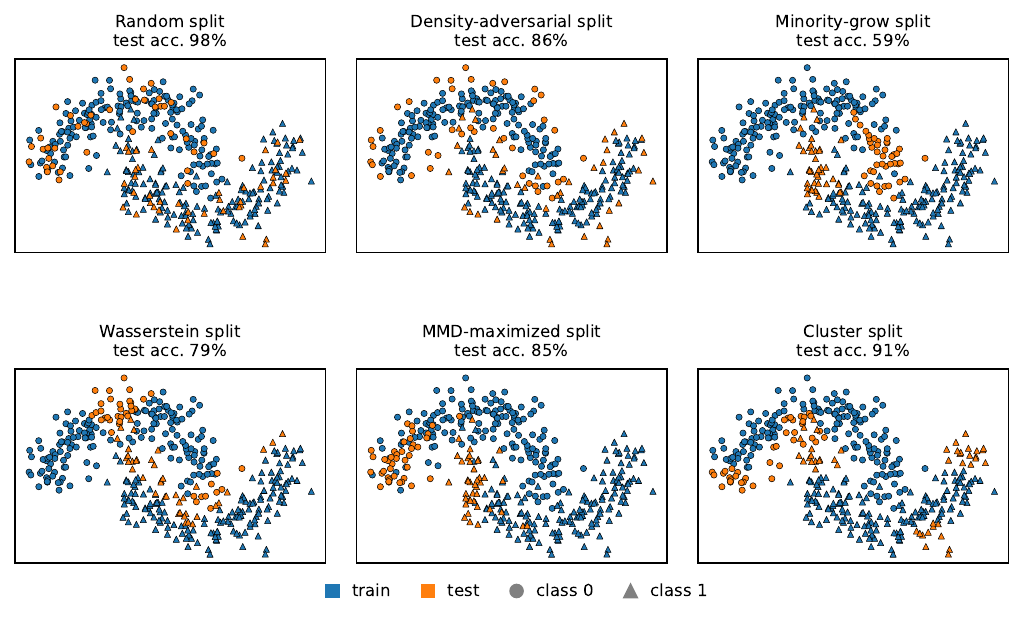}
    \caption{Six train/test splits of the two-moons dataset, each holding out 20\% as test with per-class (stratified) splitting.}
    \label{fig:splits}
\end{figure*}

Random splitting, which shuffles the examples of a dataset and holds out a fixed fraction of them for testing, is the default protocol for evaluating machine learning and NLP models.
Random splitting is convenient and simple, but it has a well-documented weakness: because the partition is random, the test set tends to resemble the training set.
Near-duplicate and otherwise similar examples can appear in both splits, and a model can obtain a high score by interpolating between training examples that are similar to the test examples.
As a result, benchmark scores can overstate how well a model generalizes to new data \citep{lewis-etal-2021-question, elangovan-etal-2021-memorization, lee-etal-2022-deduplicating}.

Prior work has constructed \emph{non-random} splits that are designed to be harder than a random partition.
\citet{sogaard-etal-2021-need} argue that random splits should not be the default and propose adversarial splits that maximize the distance between the training and test sets.
Other authors cluster the data and hold out whole clusters \citep{wecker-etal-2020-clusterdatasplit, zufle-etal-2023-latent}, order examples by a measure of difficulty such as language-model likelihood \citep{godbole-jia-2023-benchmarking} or sequence length \citep{varis-bojar-2021-sequence} and test on the tail, or build heuristic splits from modality-specific features \citep{liu-etal-2023-investigating}.
These methods consistently reveal generalization gaps that random splits do not show.
However, each method is typically introduced for a single dataset or modality, implemented for one paper, and not packaged for reuse, so that applying the method to a new dataset requires reimplementing it from scratch.

We present \texttt{splytters}, an open-source Python library that implements these methods with a single, uniform interface.
Each splitter takes an array of embeddings, which can come from text, images, audio, or tabular data, and returns the indices of the training and test examples.
\texttt{splytters} provides 31 splitters for three evaluation objectives: \emph{adversarial} splits minimize train/test similarity for hard evaluation, \emph{overlap} splits maximize train/test similarity for sanity checks and optimistic upper bounds, and \emph{balanced} splits match the train and test distributions.
Figure~\ref{fig:splits} shows that the choice of split changes the measured accuracy of the same model: on a two-moons dataset, an SVM classifier achieves a test accuracy of 98\% on a random split and between 59\% and 91\% on five adversarial splits.
Some splitters also use the class labels, for example to keep the class proportions of the two splits equal.
The library also includes \emph{sorters} that rank examples by interpretable difficulty for curriculum-style splits, and utilities that measure how adversarial or balanced a split is, such as the maximum mean discrepancy (MMD) \citep{gretton-2012-kernel} and energy distance \citep{szekely-rizzo-2013-energy} between the training and test sets.

The rest of this paper is organized as follows.
Section~\ref{sec:related} reviews related work on data splitting and harder evaluation.
Section~\ref{sec:components} describes the splitting methods and sorters that \texttt{splytters} implements.
Section~\ref{sec:usage} shows the API with short Python examples.
Section~\ref{sec:experiments} measures the effect of several splitters on the accuracy of models for text classification, regression, and named entity recognition.

\section{Related Work}
\label{sec:related}

\paragraph{Reliability of train/test splits.}
\citet{gorman-bedrick-2019-need} show that fixed ``standard'' splits produce misleading system rankings and advocate randomized resplitting, whereas \citet{sogaard-etal-2021-need} argue that random splits are too easy and propose adversarial and heuristic splits that maximize train/test dissimilarity.
Related work finds that models exploit superficial cues and train/test overlap without learning the intended capability \citep{geirhos-2020-shortcut}, which inflates benchmark scores when similar examples appear in both splits
\citep{lewis-etal-2021-question, elangovan-etal-2021-memorization, lee-etal-2022-deduplicating, larson-leach-2026-data, larson-leach-2026-measuring}.

\paragraph{Harder evaluation.}
Other work constructs evaluations that are harder by design.
Adversarial filtering iteratively removes examples that can be solved from spurious features, which yields a harder subset \citep{lebras-2020-aflite}.
Dataset cartography uses training dynamics to assign examples to easy, hard, and ambiguous regions \citep{swayamdipta-2020-cartography}.
Contrast sets perturb test instances across decision boundaries \citep{gardner-2020-contrast}, and benchmarks such as WILDS partition data by domain to measure robustness to distribution shift \citep{koh-2021-wilds}.
Like these methods, \texttt{splytters} is designed for harder evaluation, and it takes a data-centric approach: it constructs splits and difficulty rankings from the data, independently of the model that is evaluated.

\paragraph{Splitting in cheminformatics.}
Random splits also overestimate generalization in cheminformatics, because molecular datasets contain many close structural analogs, whereas the practical task is prediction on novel chemistry.
Practitioners use \emph{scaffold splits}, which group molecules by their Bemis--Murcko core structure \citep{bemis-murcko-1996-frameworks} and became the standard benchmark protocol through MoleculeNet \citep{wu-2018-moleculenet}, and \emph{time splits}, which train on past compounds and test on future compounds \citep{sheridan-2013-time}.
Because molecules with different scaffolds are often similar, \citet{guo-2024-scaffold} recommend clustering-based splits, which separate examples in embedding space as \texttt{cluster\_split} does in \texttt{splytters}.

\section{Core Components}
\label{sec:components}

This section describes the splitting methods in \texttt{splytters} (Section~\ref{sec:splitting}) and the sorters, which rank examples by difficulty (Section~\ref{sec:sorters}).

\subsection{Splitting Methods}
\label{sec:splitting}

Table~\ref{tab:splitters} lists all 31 splitters, grouped by the three objectives (\emph{adversarial}, \emph{overlap}, and \emph{balanced}), with a random baseline and group-aware variants.
The splitters share the same \texttt{(embeddings)} $\rightarrow$ \texttt{(train, test)} interface, and some of them also take class labels or group ids.
We describe several representative splitters below.

\begin{table*}[t]
\footnotesize
\setlength{\tabcolsep}{3pt}
\begin{minipage}[t]{0.485\textwidth}
\begin{tabularx}{\linewidth}{@{}l>{\raggedright\arraybackslash}X@{}}
\toprule
\textbf{Splitter} & \textbf{What it does} \\
\midrule
\multicolumn{2}{@{}l}{\emph{Adversarial}: minimize similarity (hard evaluation)} \\
\texttt{cluster}$^\dagger$ & Whole clusters to test \\
\texttt{centroid\_adversarial} & Clusters far from centroid to test \\
\texttt{distance\_adversarial} & Points far from centroid to test \\
\texttt{density\_adversarial} & Low-density points to test \\
\texttt{outlier\_adversarial} & Isolation Forest outliers to test \\
\texttt{min\_cut} & Minimum graph cut \\
\texttt{normalized\_cut} & Normalized graph cut \\
\texttt{wasserstein\_adversarial}$^\dagger$ & Farthest region (Wasserstein) \\
\texttt{mmd\_maximized}$^\dagger$ & Maximize train/test MMD \\
\texttt{minority}$^\dagger$ & Cluster-minority labels to test \\
\texttt{minority\_grow}$^\dagger$ & Minority seed grown to size \\
\texttt{class\_boundary} & Points near other classes to test \\
\texttt{decision\_boundary} & Near classifier boundary to test \\
\texttt{maximin} & Spread-out test set \\
\midrule
\multicolumn{2}{@{}l}{\emph{Baseline and grouped}} \\
\texttt{random} & Random split (baseline) \\
\texttt{group} & Each group in one split \\
\texttt{deduplicated} & Near-duplicates in one split \\
\bottomrule
\end{tabularx}
\end{minipage}\hfill
\begin{minipage}[t]{0.485\textwidth}
\begin{tabularx}{\linewidth}{@{}l>{\raggedright\arraybackslash}X@{}}
\toprule
\textbf{Splitter} & \textbf{What it does} \\
\midrule
\multicolumn{2}{@{}l}{\emph{Overlap}: maximize similarity (sanity checks)} \\
\texttt{cluster\_leak} & Split each cluster proportionally \\
\texttt{neighbor\_coverage} & $\geq k$ train neighbors per test point \\
\texttt{centroid\_matched} & Match train and test centroids \\
\texttt{stratified\_similarity} & Stratify by distance to centroid \\
\texttt{nearest\_neighbor} & Nearest neighbors kept in train \\
\texttt{central} & Closest-to-centroid points to test \\
\texttt{duplicate\_spread} & Near-duplicates in both splits \\
\texttt{max\_coverage} & Maximize test coverage by train \\
\midrule
\multicolumn{2}{@{}l}{\emph{Balanced}: match distributions (fair evaluation)} \\
\texttt{distribution\_matched} & Minimize per-feature divergence \\
\texttt{moment\_matched} & Match mean (and variance) \\
\texttt{histogram\_matched} & Match per-dimension histograms \\
\texttt{stratified\_random} & Label-stratified random split \\
\texttt{density\_balanced} & Balance local density \\
\texttt{mmd\_minimized} & Minimize train/test MMD \\
\bottomrule
\end{tabularx}
\end{minipage}
\caption{The 31 splitters in \texttt{splytters}, grouped by objective. Names omit the \texttt{\_split} suffix.
$^\dagger$ marks splitters that implement or adapt a method from earlier work: \texttt{cluster} \citep{wecker-etal-2020-clusterdatasplit, zufle-etal-2023-latent}, \texttt{wasserstein\_adversarial} \citep{sogaard-etal-2021-need}, \texttt{mmd\_maximized} \citep{napoli-white-2025-clustering}, and \texttt{minority} and \texttt{minority\_grow} \citep{reif-schwartz-2023-fighting}.
\texttt{minority\_grow} is our extension of the bias-amplified split of \citet{reif-schwartz-2023-fighting}.}
\label{tab:splitters}
\end{table*}

\paragraph{Cluster splits.}
\texttt{cluster\_split} clusters the embeddings and assigns each whole cluster to one split, so that the examples of a cluster never appear in both splits \citep{wecker-etal-2020-clusterdatasplit, zufle-etal-2023-latent}.
Its \texttt{subset\_sum} and \texttt{closest} strategies reproduce the SUBSET-SUM-SPLIT and CLOSEST-SPLIT of \citet{zufle-etal-2023-latent}, including their search over $k \in [3, 50]$ clusters (\texttt{cluster\_range=(3, 50)}).
SUBSET-SUM-SPLIT solves an exact class-balanced subset sum over clusters, and CLOSEST-SPLIT builds a test region by single linkage and fills it to the target size with individual examples.

\paragraph{Distance and density splits.}
\texttt{mmd\_maximized\_split} approximates the max-MMD objective \citep{gretton-2012-kernel} with swap-based optimization and---with \texttt{method="kernel\_kmeans"}---implements the constrained kernel $k$-means of \citet{napoli-white-2025-clustering}, whose linear-programming assignment step can also enforce per-label and per-group balance.
\texttt{density\_adversarial\_split} assigns isolated, low-density points to the test set (Figure~\ref{fig:splits}), \texttt{outlier\_adversarial\_split} assigns the outliers found by Isolation Forest \citep{liu-2008-isolation}, and \texttt{distance\_adversarial\_split} and \texttt{centroid\_adversarial\_split} assign the examples or the whole clusters that are farthest from the data centroid.
\texttt{maximin\_split} builds a maximally spread-out test set by farthest-point ($k$-center) traversal \citep{gonzalez-1985-clustering}.

\paragraph{Graph-cut splits.}
\texttt{min\_cut\_split} finds a spectral graph cut \citep{fiedler-1973-algebraic}, or optionally a Stoer--Wagner minimum cut \citep{stoer-wagner-1997-simple}, that minimizes the similarity between the two splits.
\texttt{normalized\_cut\_split} orders the examples with the spectral relaxation of the normalized graph cut \citep{shi-malik-2000-normalized} and divides this order at the requested training size.

\paragraph{Bias-amplified splits.}
\citet{reif-schwartz-2023-fighting} amplify dataset biases, which are superficial cues that a model can use to predict the label, in the training set, and test on examples for which these cues are misleading.
Their method clusters the data, assigns the examples with the majority label of each cluster to the training set, and assigns the other (``anti-biased'') examples to the test set, so that the association between cluster and label is stronger in the training set than in the full dataset.
The test set is often very small, because its size depends on the number of anti-biased examples.
Our \emph{minority-grow} split (Figure~\ref{fig:splits}) initializes the test set with the anti-biased examples and adds the closest remaining examples until the test set reaches a specified size, optionally under per-class quotas.

\paragraph{Overlap and balanced splits.}
The \emph{overlap} splitters \emph{maximize} train/test similarity, which produces optimistic upper bounds and sanity checks for leakage.
For example, \texttt{nearest\_neighbor\_split} guarantees that the nearest neighbor of each test point is in the training set, and \texttt{cluster\_leak\_split} splits each cluster proportionally so that similar examples appear in both splits.
The \emph{balanced} splitters match the train and test distributions for fair evaluation: \texttt{distribution\_matched\_split} minimizes per-feature distributional divergence, and \texttt{stratified\_random\_split} preserves label proportions.

\subsection{Sorters}
\label{sec:sorters}

A \emph{sorter} ranks samples by an interpretable difficulty or quality metric.
A sort-based split uses this ranking to hold out the highest-scoring examples as a hard test set, and the ranking can also define a curriculum ordered by difficulty.
\texttt{splytters} includes 50 sorters for embeddings, text, images, audio, and tabular data (Table~\ref{tab:sorters}).

For text, \texttt{perplexity\_score} orders examples by language-model perplexity so that the least likely examples can be held out \citep{godbole-jia-2023-benchmarking, chong-etal-2022-detecting}.
The sort-based \texttt{likelihood\_split}, which is built on this sorter, reproduces the Likelihood Splits protocol of \citet{godbole-jia-2023-benchmarking}: it ranks texts by total log-likelihood, optionally within length buckets, and holds out the least likely examples.
\texttt{character\_length} reproduces the length-threshold heuristic split of \citet{sogaard-etal-2021-need} by holding out the longest examples in order to test generalization to longer inputs \citep{varis-bojar-2021-sequence, lake-baroni-2018-generalization}.
For embeddings, \texttt{distance\_to\_mean} ranks samples by distance to the mean representation \citep{larson-etal-2019-outlier}.
The \texttt{outlier\_score} sorters for embeddings and tabular data rank samples with an outlier detector, either Isolation Forest \citep{liu-2008-isolation} or local outlier factor \citep{breunig-2000-lof}, and can detect outliers that are close to the mean.

\begin{table}[t]
\centering
\small
\resizebox{\columnwidth}{!}{%
\begin{tabular}{@{}llr@{}}
\toprule
\textbf{Modality} & \textbf{Example metrics} & \textbf{\#} \\
\midrule
Embedding & distance-to-mean, local density, outlier score & 6 \\
Text & perplexity, readability, lexical diversity, length & 8 \\
Image & brightness, contrast, color variance, compression & 7 \\
Audio & tempo, MFCC statistics, spectral centroid, duration & 17 \\
Tabular & column $z$-score, feature entropy, rarity, sparsity & 12 \\
\bottomrule
\end{tabular}%
}
\caption{Example sorter methods available in the \texttt{splytters} library.
}
\label{tab:sorters}
\end{table}

\section{Example Use}
\label{sec:usage}

The core of \texttt{splytters} depends only on \texttt{numpy}, \texttt{scipy}, and \texttt{scikit-learn}.
Integrations with heavier libraries are imported lazily, so the package installs and runs without those libraries.
\texttt{splytters} can be installed from PyPI with:
\begin{lstlisting}[language={}]
pip install splytters
\end{lstlisting}

\paragraph{Splitting.}
Each splitter is a function that takes an embeddings array of shape $(n, d)$ (and, for some splitters, labels or group ids) and returns two integer index arrays, the train and test indices:
\begin{lstlisting}
from splytters import (random_split,
    minority_grow_split)

# random baseline
tr, te = random_split(X, train_size=0.8)

# hard, class-balanced split
tr, te = minority_grow_split(
    X, y, train_size=0.8, stratify="per_class")
\end{lstlisting}
Because the splitters share this interface, they can split any data that can be embedded or represented as numeric vectors, including text, images, audio, and tabular data.
The vectors can be the output of a neural encoder or simpler features, such as bag-of-words counts.

\paragraph{Sorting.}
Sorters rank samples by an interpretable difficulty or quality metric and return a list of \texttt{(index, score)} pairs in ascending order.
The ranking can define a curriculum or a test set of the hardest examples:
\begin{lstlisting}
from splytters.sorters import distance_to_mean

ranking = distance_to_mean(X)
hardest = [i for i, _ in ranking[-50:]]
\end{lstlisting}

\paragraph{scikit-learn.}
A compatibility layer allows any splitter to be used in scikit-learn workflows.
\texttt{SplytterSplit} is a cross-validator that can be passed to \texttt{cv=}, and the \texttt{*\_train\_test\_split} helpers can replace scikit-learn's \texttt{train\_test\_split}:
\begin{lstlisting}
from sklearn.model_selection \
    import cross_validate
from splytters.sklearn_api import (
    SplytterSplit,
    adversarial_train_test_split)

cv = SplytterSplit(embeddings=X)
cross_validate(clf, X, y, cv=cv)

Xtr, Xte, ytr, yte = \
    adversarial_train_test_split(X, y)
\end{lstlisting}
For cluster-based cross-validation, \texttt{cluster\_kfold} returns a fold id for each sample so that each held-out fold contains examples from the same clusters and is balanced by label.
With \texttt{method="sds\_kmeans"}, it implements the size- and distribution-sensitive $k$-means of \citet{wecker-etal-2020-clusterdatasplit}.
The fold ids can be passed to scikit-learn's \texttt{PredefinedSplit}:
\begin{lstlisting}
from sklearn.model_selection \
    import PredefinedSplit
from splytters import cluster_kfold

folds = cluster_kfold(
    X, y, method="sds_kmeans")
cross_validate(clf, X, y,
    cv=PredefinedSplit(folds))
\end{lstlisting}

\paragraph{PyTorch and other frameworks.}
The \texttt{interop} module splits framework-native containers directly.
\texttt{to\_torch\_subsets} returns two \texttt{torch.utils.data.Subset} views that can be passed to a \texttt{DataLoader}:
\begin{lstlisting}
from splytters.interop import (
    to_torch_subsets)

train_ds, test_ds = to_torch_subsets(
        dataset, X, train_size=0.8)
\end{lstlisting}
A \texttt{splitter=} argument selects any splitter (the default is \texttt{cluster\_split}).
The helpers \texttt{split\_dataframe} and \texttt{split\_dataset} partition a pandas \texttt{DataFrame} and a HuggingFace \texttt{datasets.Dataset} in the same way.

\section{Experiments}
\label{sec:experiments}

\begin{table*}[t]
\centering
\small
\begin{tabular}{@{}lrrrrrrrr@{}}
\toprule
& \multicolumn{6}{c}{\textbf{Classification}} & \textbf{Regr.} & \textbf{NER} \\
\cmidrule(lr){2-7}\cmidrule(lr){8-8}\cmidrule(l){9-9}
\textbf{Splitter} & \textbf{TREC} & \textbf{SST-2} & \textbf{AG News} & \textbf{CLINC-150} & \textbf{DBpedia} & \textbf{Rank} & \textbf{STS-B} & \textbf{CoNLL} \\
\midrule
\textit{Random} & 87.2 & 89.3 & 89.5 & 97.7 & 97.8 & -- & 0.63 & 72.1 \\
\midrule
\multicolumn{9}{@{}l}{\emph{Feature-based adversarial}} \\
\texttt{distance\_adversarial} & 35.5 & \textbf{29.1} & 25.8 & 8.4 & 8.5 & 2.4 & \textbf{1.49} & \textbf{14.1} \\
\texttt{min\_cut} & \textbf{36.1} & 0.0 & 32.7 & 12.9 & 12.1 & 2.8 & 1.06 & 1.0 \\
\texttt{density\_adversarial} & 20.3 & 13.2 & 15.0 & 7.5 & 6.6 & 4.0 & 1.06 & 9.5 \\
\texttt{cluster} & 13.4 & 6.8 & 13.4 & 6.2 & 4.5 & 5.8 & 1.06 & 11.6 \\
\texttt{outlier\_adversarial} & 12.4 & 6.3 & 6.2 & 7.2 & 3.6 & 6.4 & 1.07 & 8.5 \\
\texttt{mmd\_maximized} & 5.0 & 0.0 & 0.5 & 4.3 & 1.1 & 8.0 & 1.00 & -0.2 \\
\midrule
\multicolumn{9}{@{}l}{\emph{Label-aware adversarial}} \\
\texttt{minority\_grow} & 34.1 & 10.5 & \textbf{33.2} & \textbf{18.8} & \textbf{15.9} & 2.0 & -- & -- \\
\texttt{decision\_boundary} & 15.4 & 16.7 & 15.5 & 4.8 & 4.6 & 4.6 & -- & -- \\
\midrule
\multicolumn{9}{@{}l}{\emph{Sort-based}} \\
target-sorted & -- & -- & -- & -- & -- & -- & 1.38 & -- \\
\midrule
\multicolumn{9}{@{}l}{\emph{Controls}} \\
\texttt{stratified\_random} & -0.5 & -0.3 & -0.3 & -0.1 & 0.1 & 9.4 & -- & -- \\
\texttt{cluster\_leak} & 0.2 & -0.3 & -0.3 & -0.3 & 0.1 & 9.6 & 1.00 & -0.3 \\
\bottomrule
\end{tabular}
\caption{Effect of each splitter relative to \texttt{random\_split}: the drop in balanced accuracy (classification) and entity-level F1 (NER), in percentage points, and the ratio of mean absolute errors (regression). The first row shows the score on \texttt{random\_split}. Higher values mean harder test splits, bold marks the hardest, and Rank is the average over the classification datasets.}
\label{tab:text-results}
\end{table*}

We measure the effect of the splitters on three types of NLP tasks: text classification, regression, and named entity recognition (NER).
For classification, we use TREC \citep{li-roth-2002-learning}, SST-2 \citep{socher-etal-2013-recursive}, AG News and DBpedia \citep{zhang-etal-2015-character}, and the in-scope intents of CLINC-150 \citep{larson-etal-2019-evaluation}, which have between 2 and 150 classes.
For regression, we predict the similarity scores of the sentence pairs of STS-B \citep{cer-etal-2017-semeval}, and for NER we use CoNLL-2003 \citep{tjong-kim-sang-de-meulder-2003-introduction}.

\paragraph{Setup.}
We represent the text with the frozen \texttt{all-mpnet-base-v2} sentence encoder \citep{reimers-gurevych-2019-sentence, song-etal-2020-mpnet}, and train a linear SVM for classification, ridge regression for STS-B, and logistic regression on the contextual token embeddings of the encoder for NER, where the splitters assign whole sentences to the splits.
We report the difference from \texttt{random\_split} in balanced accuracy, mean absolute error (MAE), and entity-level F1, averaged over 10 seeds with 70\% of the examples in the training split.
We apply the splitters that do not use labels to each class separately for classification, with \texttt{per\_class\_split}, and to the whole dataset for STS-B and CoNLL-2003, where the label-aware splitters do not apply because the data samples do not have sentence-level class labels.
For STS-B, a target-sorted split, built with \texttt{sorted\_stratified\_split}, assigns the pairs with the highest similarity scores to the test split.
We use \texttt{stratified\_random\_split} and \texttt{cluster\_leak\_split} as controls, for which we expect scores close to those of \texttt{random\_split}.
Appendices~\ref{app:full} and~\ref{app:noncls} describe the datasets and the setup in detail, and report results on image and audio tasks.

\paragraph{Results.}
Table~\ref{tab:text-results} shows that the adversarial splitters reduce balanced accuracy by up to 36.1 points, increase the MAE on STS-B by a factor of up to 1.49 and reduce the F1 on CoNLL-2003 by up to 14.1 points.
We observe minimal changes under the \texttt{stratified\_random\_split} and \texttt{cluster\_leak\_split} controls: at most 0.5 points in balanced accuracy, less than 0.1\% in MAE and 0.3 points in F1.
Among the classification datasets, \texttt{minority\_grow\_split} is the hardest splitter on AG News, CLINC-150, and DBpedia, \texttt{min\_cut\_split} on TREC, and \texttt{distance\_adversarial\_split} on SST-2. We also find that \texttt{distance\_adversarial\_split} is the hardest on both STS-B and CoNLL-2003.
The ranking of the eight adversarial splitters is similar across the five classification datasets, with a Kendall's coefficient of concordance of $W = 0.75$ \citep{kendall-babington-smith-1939-problem}.
The effect of a splitter can differ between tasks: \texttt{min\_cut\_split} has the third-best average rank on classification, but reduces F1 by only 1.0 point on CoNLL-2003.

A dissimilar test split is not always a harder one.
On SST-2, a classifier trained to distinguish the training split of \texttt{min\_cut\_split} from its test split has an AUC of 0.995, but the drop in accuracy is 0.0 points, 
because the training and test sentences differ in length (7.2 and 15.3 words on average), and the SVM is about equally accurate on short and long sentences.
We therefore recommend using \texttt{split\_report}, which measures how dissimilar the two splits are, together with the accuracy of the model on each split.

\paragraph{Fine-tuned models.}
We also fine-tune two models on splits of the 400,000 documents of RVL-CDIP \citep{harley-etal-2015-evaluation}, built from CLIP embeddings of the page images: LayoutLMv3 \citep{huang-etal-2022-layoutlmv3}, which uses the text, layout, and image of each page, and ResNet-50 \citep{he-etal-2016-deep}, which uses only the image.
Table~\ref{tab:rvlcdip} shows that the two adversarial splitters reduce balanced accuracy by 11.9 to 24.3 points, whereas the two overlap splitters reduce it by at most 0.2 points or increase it (details in Appendix~\ref{app:rvlcdip}).

\begin{table}[t]
\centering
\small
\begin{tabular}{@{}lrr@{}}
\toprule
\textbf{Splitter} & \textbf{ResNet-50} & \textbf{LayoutLMv3} \\
\midrule
\textit{Random (accuracy)} & 90.6 & 94.9 \\
\midrule
\texttt{distance\_adversarial} & 19.9 & 11.9 \\
\texttt{cluster} & 24.3 & 14.9 \\
\midrule
\texttt{nearest\_neighbor} & 0.0 & 0.2 \\
\texttt{central} & -4.1 & -2.1 \\
\bottomrule
\end{tabular}
\caption{Drop in balanced accuracy (points) on RVL-CDIP relative to \texttt{random\_split}, whose balanced accuracy (\%) is in the first row.}
\label{tab:rvlcdip}
\end{table}

\section{Conclusion}

We presented \texttt{splytters}, an open-source Python library with 31 splitters for adversarial, overlap, and balanced evaluation and 50 sorters that rank examples by difficulty.
In our experiments, the adversarial splitters reduce balanced accuracy by up to 36.1 points on text classification, increase the mean absolute error of text regression by a factor of up to 1.49, and reduce the F1 of NER by up to 14.1 points, and the hardest splitter differs between datasets and tasks, so we recommend evaluating with more than one splitter.
\texttt{splytters} is available under the MIT license at \url{https://github.com/gxlarson/splytters}, with documentation at \url{https://splytters.readthedocs.io}.

\section*{Limitations}

Adversarial splitters assign atypical examples --- such as outliers, examples in low-density regions, and the minority examples of a cluster --- to the test split.
These examples can correspond to underrepresented groups, dialects, or document types.
An adversarial test split can therefore measure the accuracy of a model on these groups, but a model trained on the corresponding training split has seen fewer of them, so a model intended for deployment should be trained on all of the data.

The splitters measure similarity in the embedding space of an encoder, so the biases of the encoder determine which examples are kept apart, and the same splitter can produce different splits with a different encoder.

A choice among 31 splitters also makes it easy to select a split on which a model looks worse or, with an overlap splitter, better.
We recommend reporting results on \texttt{random\_split} alongside any other split, together with the splitter, its parameters, and the seeds, and using the overlap splitters only as sanity checks.

\section*{Ethical Considerations}

An adversarial split can assign most of the examples of a demographic group to the test split, and a model trained on the corresponding training split can be less accurate for that group.
The adversarial splitters are intended to test for such biases before a model is used, and the resulting accuracy should be reported as a measure of robustness.
Users should check how demographic groups are distributed across the two splits, where this information is available, and report accuracy for each group.

\bibliography{custom}
\bibliographystyle{acl_natbib}

\clearpage
\appendix

\section{Results for all datasets}
\label{app:full}

\begin{table*}[t]
\centering
\small
\resizebox{\textwidth}{!}{%
\begin{tabular}{@{}lrrrrrrrrrrrrr@{}}
\toprule
& \multicolumn{4}{c}{\textbf{Image}} & \multicolumn{5}{c}{\textbf{Text}} & \multicolumn{3}{c}{\textbf{Audio}} & \\
\cmidrule(lr){2-5}\cmidrule(lr){6-10}\cmidrule(lr){11-13}
\textbf{Splitter} & \textbf{C10} & \textbf{C100} & \textbf{FMN} & \textbf{Pets} & \textbf{TREC} & \textbf{SST-2} & \textbf{AGN} & \textbf{CLINC} & \textbf{DBP} & \textbf{ESC} & \textbf{US8K} & \textbf{SC} & \textbf{Rank} \\
\midrule
\textit{Random (accuracy)} & 92.8 & 70.9 & 87.0 & 92.3 & 87.2 & 89.3 & 89.5 & 97.7 & 97.8 & 73.5 & 69.2 & 65.7 & -- \\
\midrule
\multicolumn{14}{@{}l}{\emph{Feature-based adversarial}} \\
\texttt{min\_cut} & 16.9 & \textbf{27.2} & 22.6 & 15.6 & \textbf{36.1} & 0.0 & 32.7 & 12.9 & 12.1 & 23.9 & 34.4 & 32.7 & 2.25 \\
\texttt{distance\_adversarial} & 12.1 & 23.0 & 18.0 & 11.3 & 35.5 & \textbf{29.1} & 25.8 & 8.4 & 8.5 & \textbf{25.2} & \textbf{34.9} & \textbf{36.1} & 2.50 \\
\texttt{density\_adversarial} & 12.7 & 24.2 & 14.4 & 10.5 & 20.3 & 13.2 & 15.0 & 7.5 & 6.6 & 20.3 & 20.9 & 18.4 & 4.00 \\
\texttt{cluster} & 8.9 & 13.9 & 17.0 & 8.8 & 13.4 & 6.8 & 13.4 & 6.2 & 4.5 & 15.4 & 20.0 & 18.4 & 5.67 \\
\texttt{outlier\_adversarial} & 10.2 & 18.1 & 14.1 & 8.2 & 12.4 & 6.3 & 6.2 & 7.2 & 3.6 & 10.9 & 10.0 & 10.2 & 6.67 \\
\texttt{mmd\_maximized} & 0.5 & 13.8 & 1.7 & 11.3 & 5.0 & 0.0 & 0.5 & 4.3 & 1.1 & 21.2 & 20.7 & 12.2 & 6.75 \\
\midrule
\multicolumn{14}{@{}l}{\emph{Label-aware adversarial}} \\
\texttt{minority\_grow} & \textbf{25.2} & 24.7 & \textbf{42.1} & \textbf{20.4} & 34.1 & 10.5 & \textbf{33.2} & \textbf{18.8} & \textbf{15.9} & 14.6 & 27.3 & 6.9 & 2.67 \\
\texttt{decision\_boundary (SVM)} & 12.1 & 17.4 & 14.3 & 9.2 & 15.4 & 16.7 & 15.5 & 4.8 & 4.6 & 8.5 & 12.4 & 17.1 & 5.50 \\
\texttt{decision\_boundary (LR)} & 13.2 & 21.5 & 13.6 & 11.8 & 16.2 & 17.0 & 15.3 & 5.3 & 4.1 & 11.0 & 10.8 & 16.5 & -- \\
\midrule
\multicolumn{14}{@{}l}{\emph{Controls}} \\
\texttt{stratified\_random} & 0.0 & -0.2 & 0.5 & -0.2 & -0.5 & -0.3 & -0.3 & -0.1 & 0.1 & -0.6 & 0.1 & 0.5 & 9.25 \\
\texttt{cluster\_leak} & 0.0 & -1.0 & 0.2 & -0.7 & 0.2 & -0.3 & -0.3 & -0.3 & 0.1 & -2.7 & -0.4 & -1.3 & 9.75 \\
\bottomrule
\end{tabular}%
}
\caption{Drop in balanced accuracy (points) relative to \texttt{random\_split}, averaged over 10 seeds and the encoders of each modality. The first row shows the balanced accuracy (\%) on \texttt{random\_split}, and bold marks the largest drop. C10 and C100 are CIFAR-10 and CIFAR-100, FMN is Fashion-MNIST, AGN is AG News, DBP is DBpedia, ESC is ESC-50, US8K is UrbanSound8K, and SC is Speech Commands.}
\label{tab:full-results}
\end{table*}

\paragraph{Text.}
We use all 5,952 questions of TREC, all 22,500 in-scope queries of CLINC-150, and 4,000, 2,000, and 600 examples per class of SST-2, AG News, and DBpedia.
The splitters receive a 128-dimensional PCA projection of the normalized embeddings, and the linear SVM is trained on the full embeddings.
\texttt{minority\_grow\_split} uses as many clusters as there are classes. When we vary the number of clusters from 5 to 200, its drop changes by a median of 15.9 points across the 12 datasets of Table~\ref{tab:full-results}.

\paragraph{Images.}
We use stratified samples of 12,000 training images of CIFAR-10, CIFAR-100 \citep{krizhevsky-2009-learning}, and Fashion-MNIST \citep{xiao-etal-2017-fashion}, and all 3,680 training images of Oxford-IIIT Pets \citep{parkhi-etal-2012-cats}, with CLIP ViT-B/32 \citep{radford-etal-2021-learning}, DINOv2-small \citep{oquab-etal-2024-dinov2}, and ResNet-50 \citep{he-etal-2016-deep} as frozen encoders.
The splitters receive the embeddings without a PCA projection, and the linear SVM is trained with stochastic gradient descent.

\paragraph{Audio.}
We use ESC-50 \citep{piczak-2015-esc}, UrbanSound8K \citep{salamon-etal-2014-dataset}, and Speech Commands \citep{warden-2018-speech} with AST \citep{gong-etal-2021-ast} and wav2vec~2.0 \citep{baevski-etal-2020-wav2vec} as frozen encoders, and the same PCA projection and linear SVM as for text.
In UrbanSound8K, excerpts of the same recording are near-duplicates, so we keep one excerpt per recording (1,297 of 8,732).
We sample 11,999 clips of Speech Commands. One class has a single clip, so we add a second one for \texttt{decision\_boundary\_split} and \texttt{stratified\_random\_split}, which require two examples per class.
The AST embeddings of Speech Commands are averages over all frames, including padding, and have a balanced accuracy of 55.2\% on \texttt{random\_split}, against 76.1\% for wav2vec~2.0, whose embeddings exclude the padding, so the results with AST on this dataset should be interpreted with caution.

\paragraph{Results.}
Table~\ref{tab:full-results} shows the results for the 12 datasets, with ranks computed from the unrounded drops.
Three splitters have similar average ranks: \texttt{min\_cut\_split} (2.25), \texttt{distance\_adversarial\_split} (2.50) and \texttt{minority\_grow\_split} (2.67).
The ranking of the adversarial splitters is similar across the datasets of each modality ($W = 0.78$ for both images and audio), and the controls differ from \texttt{random\_split} by at most 2.7 points.
When the similarity graph of a class is disconnected, \texttt{min\_cut\_split} assigns the smallest components to the test split first. If it assigns the largest first, its drop is lower by more than 0.5 points on 15 of the 23 combinations of dataset and encoder, but its average rank is the same.
Table~\ref{tab:full-results} also shows \texttt{decision\_boundary\_split} with a logistic regression model (LR), which we do not rank.

\section{Regression, sequence labeling, and segmentation}
\label{app:noncls}

We run every splitter on the whole set of examples without labels, with 70\% of the examples in the training split and 10 seeds.
We report the mean absolute error for regression because the Spearman correlation also depends on the spread of the targets in the test split. For example, the test splits of \texttt{min\_cut\_split} on UTKFace have a smaller spread of ages than those of \texttt{random\_split} (standard deviations of 14.5 and 23.9 years), and the correlation drops by 46.8 points.

\begin{table}[t]
\centering
\small
\begin{tabular}{@{}lrr@{}}
\toprule
\textbf{Splitter} & \textbf{UTKFace} & \textbf{VOC} \\
\midrule
\textit{Random} & 7.1 & 75.5 \\
\midrule
\texttt{distance\_adversarial} & 1.11 & 9.0 \\
\texttt{density\_adversarial} & 1.25 & 21.2 \\
\texttt{outlier\_adversarial} & 1.12 & 0.2 \\
\texttt{cluster} & 1.26 & \textbf{52.3} \\
\texttt{min\_cut} & 1.47 & 24.2 \\
\texttt{mmd\_maximized} & 1.00 & 0.3 \\
target-sorted & \textbf{4.07} & -- \\
\texttt{cluster\_leak} & 1.00 & -0.5 \\
\bottomrule
\end{tabular}
\caption{Ratio of mean absolute errors (UTKFace) and drop in mIoU in points (VOC), relative to \texttt{random\_split}, whose mean absolute error (years) and mIoU (\%) are in the first row. Bold marks the hardest split.}
\label{tab:noncls}
\end{table}

\paragraph{STS-B.}
We use the 7,249 sentence pairs of the GLUE training and validation splits, with similarity scores from 0 to 5.
Each pair is represented by $[u, v, |u - v|]$, where $u$ and $v$ are the embeddings of the two sentences, and the splitters receive a 128-dimensional PCA projection of this vector.

\paragraph{CoNLL-2003.}
We sample 8,000 of the 20,744 sentences.
The splitters receive a 128-dimensional PCA projection of the mean-pooled embedding of each sentence.
The logistic regression model receives, for each word, the contextual embedding of its first subword token from the last layer of the encoder.

\paragraph{UTKFace.}
We use the 7,623 training images of UTK-Face-revised, a subset of UTKFace \citep{zhang-etal-2017-age} with ages from 1 to 110 years.
A ridge regression model predicts the age from the same three image encoders as in Appendix~\ref{app:full}, and Table~\ref{tab:noncls} records the mean over the encoders.

\paragraph{Pascal VOC 2012.}
We use the 1,464 training images with segmentation masks \citep{everingham-etal-2010-pascal}, resized and center-cropped to 224 by 224 pixels together with their masks.
The splitters receive the DINOv2-small embedding of the class token of each image, and a logistic regression model predicts the class of each of the 16 by 16 patches from its embedding. We report the mean intersection over union (mIoU) over these patches.
Because the images do not have a single class label, the splits can change the shares of the object classes. With \texttt{cluster\_split} and \texttt{min\_cut\_split}, 70\% and 41\% of the object patches in the test split belong to classes with less than 0.5\% of the object patches in the training split, which explains part of their drops.

\section{Fine-tuning on RVL-CDIP}
\label{app:rvlcdip}

We apply the splitters to frozen CLIP ViT-B/32 embeddings of the page images, separately for each of the 16 classes, and select 20\% of the documents as the test split.
We use a random 10\% of the documents as a validation split to select the checkpoint.
We train ResNet-50, initialized with ImageNet weights, for 30 epochs with SGD and a learning rate of $10^{-3}$, and LayoutLMv3-base, with the words and boxes produced by Amazon Textract, for 10 epochs with Adam and a learning rate of $10^{-5}$.
We train each model once on each split, except that we run \texttt{cluster\_split}, with 10 clusters per class and about 21.6\% of the documents in the test split, with three seeds.
Its drops range from 22.6 to 25.5 points for ResNet-50 and from 14.0 to 16.2 points for LayoutLMv3.
\texttt{cluster\_split} reduces accuracy more than \texttt{distance\_adversarial\_split} on RVL-CDIP, whereas \texttt{distance\_adversarial\_split} has the better average rank with frozen embeddings (Table~\ref{tab:full-results}).

\end{document}